\documentclass[12pt, draftclsnofoot, onecolumn]{IEEEtran}
\usepackage{cite}
\usepackage{physics}
\usepackage{amsmath, amssymb, amsfonts, amssymb, amsthm}
\usepackage{lipsum}
\usepackage{mathtools}
\usepackage{cuted}
\usepackage{algorithm}
\usepackage{algorithmicx}
\usepackage{algpseudocode}
\usepackage{graphicx}
\usepackage{epstopdf}
\usepackage{tikz}
\usepackage{bbm}
\usepackage{textcomp} 
\usepackage{adjustbox}
\usepackage{scalerel}
\usepackage{subfig}
\usepackage{xcolor}
\usepackage{stfloats}

\begin{document}

{\Large \textbf{Notice:} This work has been submitted to the IEEE for possible publication. Copyright may be transferred without notice, after which this version may no longer be accessible.}
\clearpage

\title{Role of Sensing and Computer Vision in 6G Wireless Communications}

\author{Seungnyun Kim,~\IEEEmembership{Member,~IEEE,} Jihoon Moon,~\IEEEmembership{Member,~IEEE,} Jinhong Kim,~\IEEEmembership{Member,~IEEE,} Yongjun Ahn,~\IEEEmembership{Member,~IEEE,} Donghoon Kim,~\IEEEmembership{Member,~IEEE,} Sunwoo Kim,~\IEEEmembership{Member,~IEEE,} Kyuhong Shim,~\IEEEmembership{Member,~IEEE,} and Byonghyo Shim,~\IEEEmembership{Senior Member,~IEEE}
}

\maketitle

\begin{abstract}
Recently, we are witnessing the remarkable progress and widespread adoption of sensing technologies in autonomous driving, robotics, and metaverse.
Considering the rapid advancement of computer vision (CV) technology to analyze the sensing information, we anticipate a proliferation of wireless applications exploiting the sensing and CV technologies in 6G.
In this article, we provide a holistic overview of the sensing and CV-aided wireless communications (SVWC) framework for 6G.
By analyzing the high-resolution sensing information through the powerful CV techniques, SVWC can quickly and accurately understand the wireless environments and then perform the wireless tasks.
To demonstrate the efficacy of SVWC, we design the whole process of SVWC including the sensing dataset collection, DL model training, and execution of realistic wireless tasks.
From the numerical evaluations on 6G communication scenarios, we show that SVWC achieves considerable performance gains over the conventional 5G systems in terms of positioning accuracy, data rate, and access latency.
\end{abstract}
\newpage

\section{Introduction}
Recently, there has been growing interest in immersive mobile services such as digital twin and metaverse that offer a more accurate representation of the physical world.
Since the virtual model reflecting the physical world uses high-fidelity holographic displays (e.g., 8K augmented reality (AR) streaming, 16K virtual reality (VR) streaming, and extended reality (XR)), these services require far beyond-gigabit data rates~\cite{dang2020should}.
One of the most effective ways to meet this unprecedented high data rate demand is to explore the ultra-high frequency spectrum.
Indeed, the millimeter (mmWave) band (sub-$6\,\text{GHz}$ and above $24\,\text{GHz}$) communication has been adopted in 5G New Radio (NR) and the terahertz (THz) band ($0.1\sim 10\,\text{THz}$) communication is considered as a candidate technology for 6G systems~\cite{chaccour2022seven}.

When the transmission frequency increases, communication distance is reduced due to the severe path loss and strong directivity.
To cope with the reduced communication distance, an open radio access network (O-RAN) architecture where the digital unit (DU) handles the baseband processing and the densely deployed remote radio heads (RRHs) deal with the signal transmission and reception has received much attention recently.
Two notable features of the densified network using the high frequency spectrum are: 1) the concentration of the transmit signal on the line-of-sight (LoS) path and 2) the reduced cell coverage in the order of a few tens of meters.
In line with the trend of communication area being close to the human visual area, sensing technology that observes the surrounding environments through various sensing modalities (e.g., RGB camera, depth camera, infrared camera, LiDAR, radar) has received considerable attention these days.
For example, cell phone cameras boast a pixel count of $200$ million, leading to elevated image quality and the ability to capture minute details that even the human eye cannot perceive.

Recently, we have witnessed the remarkable success and widespread adoption of the sensing technologies in autonomous driving, robotics, and unmanned air mobility (UAM).
In parallel with the advancement of sensing technology, computer vision (CV) technology that analyzes and interprets the sensor information has made a gigantic leap with an aid of deep learning (DL) technology~\cite{nishio2021wireless}.
Nowadays, we are seeing the bewildering variety of DL-based CV techniques including object detection, semantic segmentation, image captioning, and object tracking.
Considering its potential benefits and speed of advancement, one cannot pin down the dimension of CV applications for the upcoming 6G systems.

\begin{figure}[t]
    \centering
    \includegraphics[width=0.98\linewidth]{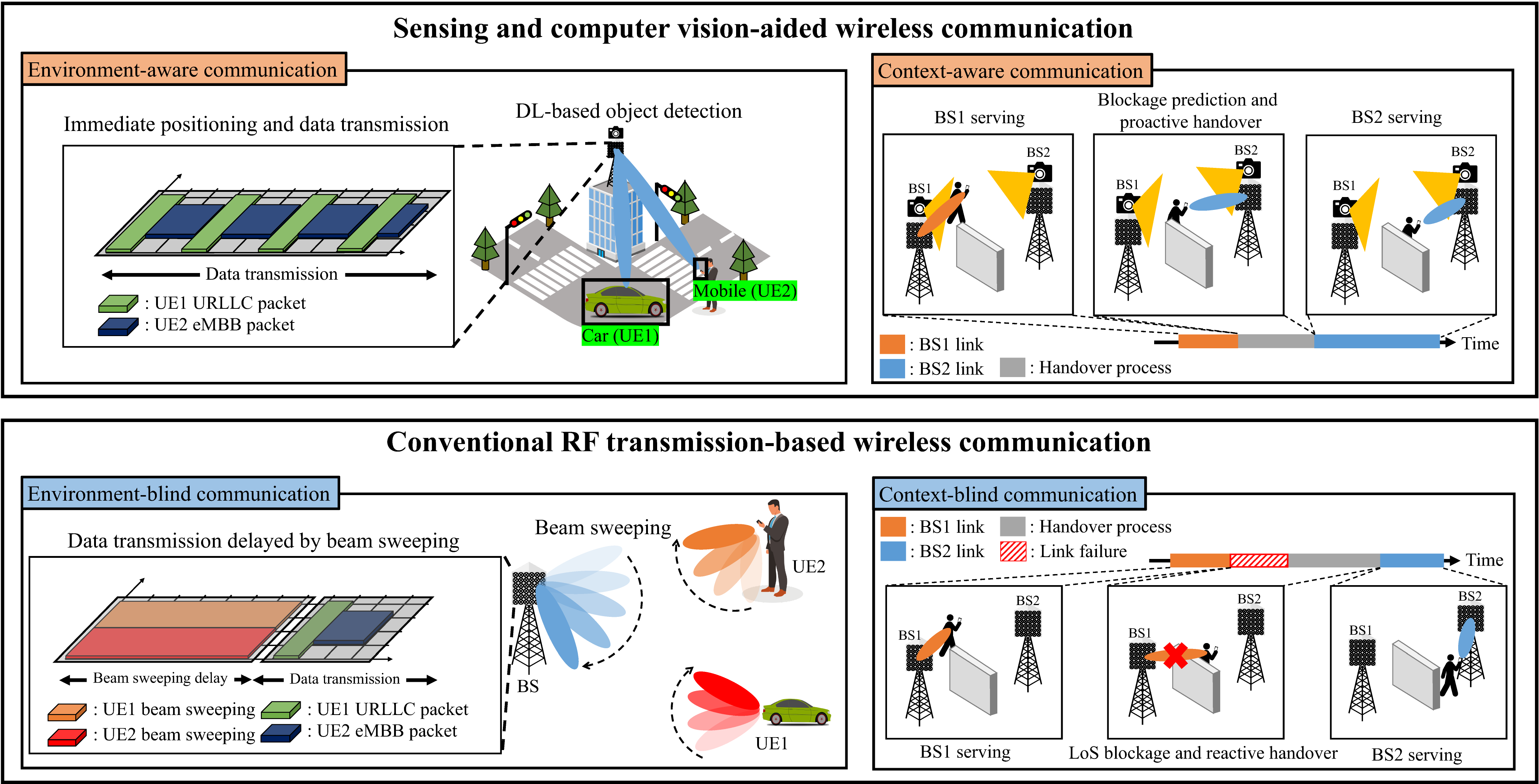}
    \caption{Comparison of conventional RF transmission-based communication system and SVWC.}
    \label{fig1}
\end{figure}

An aim of this article is to introduce a novel communication framework based on the sensing and CV technologies referred to as \textit{sensing and CV-aided wireless communications} (SVWC).
Since 6G communications will bring about a range of emerging applications, including unmanned aerial vehicles (UAV), swarm robots, and smart factories/farms, which demand not only hyper-fast data rate but also extremely-low latency and ultra-reliable positioning capabilities, sensing and CV can play a vital role in achieving ambitious goals in 6G (see Fig.~\ref{fig1}).
The major benefits of SVWC when compared to the conventional systems relying exclusively on the RF transmission, are as follows:

\begin{itemize}
\item \textbf{Environment-awareness:}
By analyzing the sensing information, various components of the wireless environment (e.g., obstacle location, channel condition, and distance between serving base station (BS) and neighboring BS) can be identified with minimal exchange of control signaling (e.g., physical uplink control channel (PUCCH) and physical downlink control channel (PDCCH)).
In doing so, SVWC can significantly reduce the transmission latency, power consumption, and resource overhead (e.g., pilot symbols and feedback bits).
Moreover, by interpreting the sensing information using DL-based CV techniques, SVWC can process high-level intelligent functionalities such as beamforming, blockage prediction, and cell association.

\item \textbf{Context-awareness:}
The high-resolution sensing data inherently contains contextual and semantic information that wireless RF signals cannot provide.
In fact, by capturing human behaviors and activities such as movement, user density, and gestures, SVWC can extract detailed and real-time information of mobiles which can be used for the context-aware services such as event (emergency) detection, gesture recognition, traffic control, predictive handover, and mobility management.

\item \textbf{Parallel processing capability:}
With the relentless growth of the number of connected devices, the overhead to perform the user identification, random access, and initial access will significantly increase in 6G era.
SVWC alleviates the problem since the sensing image can recognize multiple objects including connected devices (e.g., mobiles, tablets, and laptops), reflecting scatterers (e.g., trees and walls), and also BSs (e.g., picocells and femtocells).
By the concurrent processing of random access, scheduling, cell association, and handover, SVWC can greatly reduce the access latency and improve the resource utilization efficiency.
\end{itemize}

To demonstrate the potential benefit of SVWC, we design the whole process of SVWC including the dataset collection, DL model training, and validation.
From the experiments conducted on realistic 6G communication scenarios, we demonstrate that SVWC achieves more than $91\%$ reduction in positioning error, $29\%$ reduction in access latency, and $84\%$ improvement in data rate over the conventional 5G NR systems.

\section{Basics of Sensing and Computer Vision}
Sensing is the process to detect and capture visual, auditory, and tactile information about the physical world through the sensing device (e.g., RGB cameras, microphones, and touch sensors) and CV is the technique to perceive and comprehend the physical world by analyzing the acquired sensing information, particularly visual sensing data.
With an adoption of DL technique, performance of CV techniques has been improved dramatically, surpassing the cognitive ability of humans in face detection, medical image diagnosis, and visual surveillance~\cite{zou2023object}.
Through extensive training and validation using vision datasets (e.g., MS-COCO and ImageNet), the DL-based CV technique can extract nuanced features, recognize objects, discern patterns, and interpret environments and semantic meaning of images and videos.
Overall, the CV process consists of three major steps (see Fig.~\ref{fig2}):

\begin{enumerate}
\item In the first step, called \textit{vision acquisition}, 2D and 3D sensing information is captured using RGB cameras, LiDAR, radar, and infrared cameras.
For instance, RGB camera captures the high-resolution images with RGB pixel values and LiDAR can capture the precise depth information on the order of centimeters.
To further enhance the detection probability and also better understand wireless environments, approaches to use multi-modal sensor devices (e.g., multiple RGB cameras or RGB camera+LiDAR) can be employed.

\begin{figure}[t]
    \centering
    \includegraphics[width=1\linewidth]{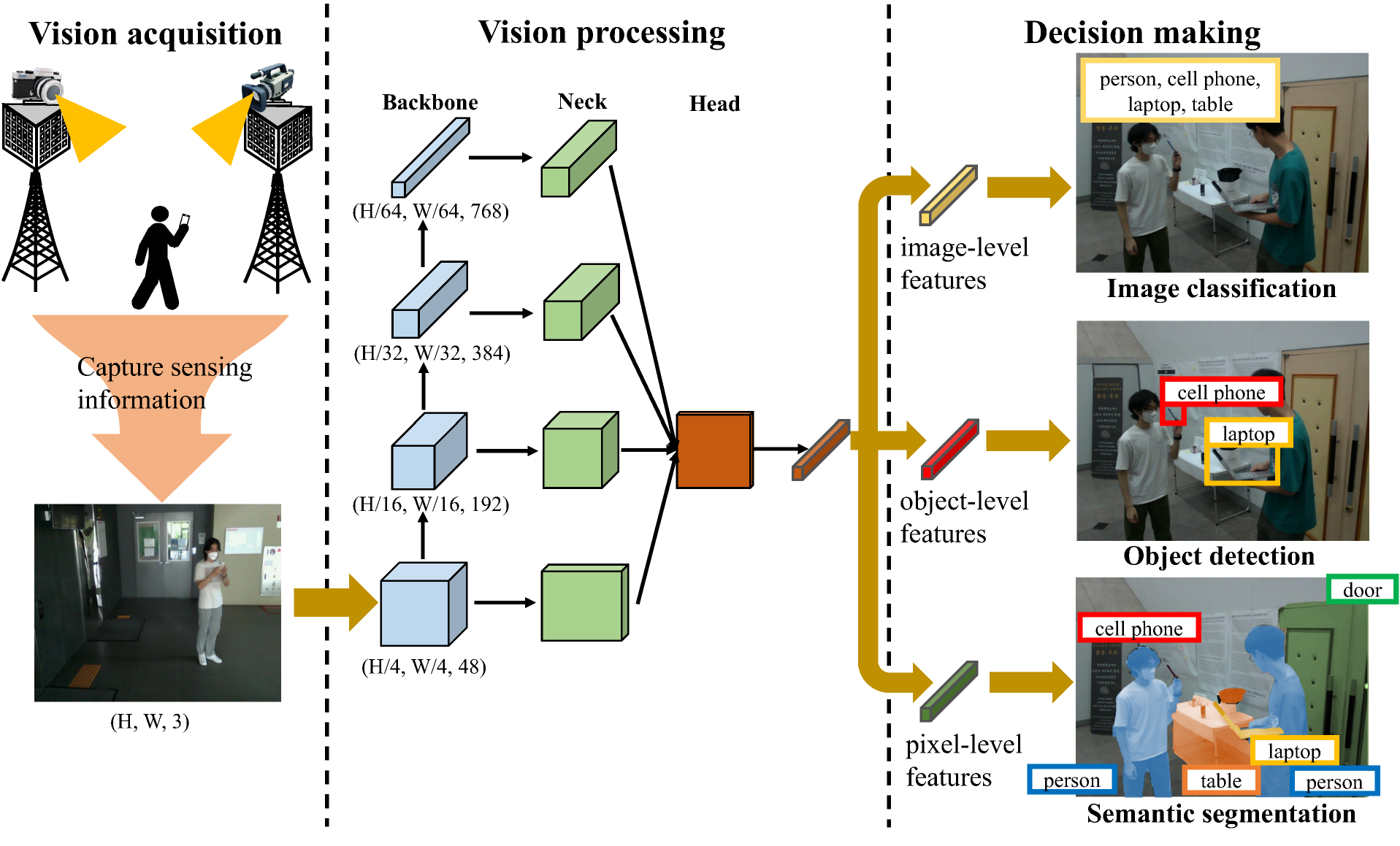}
    \caption{Illustration of CV-based image processing mechanism.}
    \label{fig2}
\end{figure}

\item In the second step, called \textit{vision processing}, a feature map, a low-dimensional vector containing core information (e.g., multiple-input multiple-output (MIMO) antenna array structure and locations of scatterers), is extracted from the acquired sensing data via the deep neural network (DNN).
One natural option to extract the spatially correlated features is the convolutional neural network (CNN) which applies multiple filter kernels to convolve the input image.
Recently, Transformer, a DL model that differentially computes the significance of input elements using the attention mechanism, has received a great deal of attention~\cite{vaswani2017attention}.
Using the attention mechanism, Transformer can extract not only the correlation of the adjacent pixels but also those of the spaced-apart pixels, thereby generating the local and global features in the sensing data simultaneously.

\item In the last step, called \textit{decision-making}, extracted features are used to complete the downstream CV task.
Overall, a CV task is divided into three major categories: 1) image classification (image-level), 2) object detection (object-level), and 3) semantic segmentation (pixel-level).
In the image classification, images are grouped into categories of interest (e.g., cat or dog).
In the object detection, locations of the real-world objects (e.g., cars, laptops, and persons) are identified in a form of the bounding box, the smallest rectangular-shaped box containing the target object.
In the semantic segmentation, a class for each pixel is identified for the pixel-wise understanding of an image (see Fig.~\ref{fig2}).
Other than these, there are many intriguing CV tasks such as image captioning, object tracking, action recognition, and motion estimation.
\end{enumerate}

\section{Sensing and Computer Vision-aided Wireless Communications}
Main strength of sensing and CV technologies is that they provide not only fast and accurate identification of wireless environments (e.g. obstacles, reflectors) and target objects (e.g., BS, mobile, vehicle, and drone) but also contextual and cognitive understanding of the interactions between the wireless environments and objects (e.g., multipath propagation, blockage, and interference).
We here list five exemplary wireless applications in which CV technology can be beneficial.

\subsection{Sensing and CV-aided Beam Management}
In the mmWave and THz communications, the beamforming technique is used to compensate for the severe path loss.
In 5G NR, a two-step beam management process has been employed to identify the beam direction.
In the first step called beam sweeping, the BS transmits the beam codewords containing the synchronization signal blocks (SSBs).
The mobile measures the reference signal power of the SSB beams and then feeds back the index of the SSB beam corresponding to the largest reference signals received power (RSRP).
In the second step called beam refinement, the BS identifies the mobile’s location by sending multiple channel state information reference signal (CSI-RS) to the direction identified by the beam sweeping.
The beam management strategy of 5G NR has two notable drawbacks: \textit{beam misalignment} caused by mobility and the use of a finite number of beam codewords (e.g., in the 5G FR2 band, up to $64$ SSB beams are transmitted) and \textit{beam training latency} caused by the time-consuming two-step handshaking process.

\begin{figure}[t]
    \centering
    \includegraphics[width=1\linewidth]{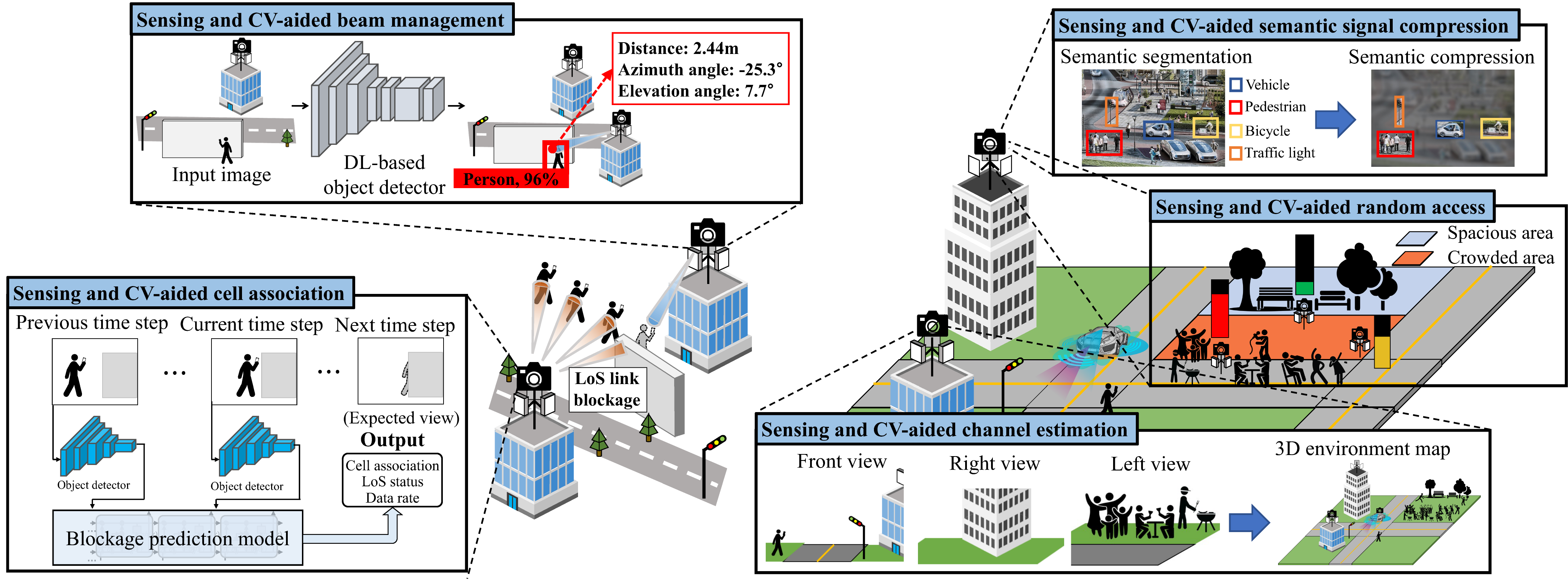}
    \caption{Illustrations of sensing and CV-aided beam management, cell association, channel estimation, semantic signal compression, and random access.}
    \label{fig3}
\end{figure}

SVWC is beneficial in achieving an accurate yet fast beamforming. 
In this approach, a DL-based object detector directly finds out the position of a mobile from the sensing information and determines its class (e.g., human, vehicle, mobile)~\cite{ahn2022towards}.
Once the position of a target mobile is acquired, the BS generates a directional beam heading toward the mobile (see Fig.~\ref{fig3}).
Since the location information is derived from the captured image, an overhead caused by the beam sweeping and refinement processes can be prevented, resulting in a considerable reduction of the beam management latency.
Also, seamless tracking of a mobile is possible via the object tracking so that the beam misalignment caused by the movement of a mobile can be reduced~\cite{charan2021vision}.

\subsection{Sensing and CV-aided Proactive Cell Association}
Due to the high directivity and severe path loss in the mmWave and THz bands, scattering and reflection of signals are negligible, leaving the LoS path as the dominant one in signal propagation.
In fact, if the LoS path is blocked in the mmWave systems, the signal power will drop severely, causing a radio link failure (RLF).
In 5G NR, the BS receives the signal power report (e.g., RSRP) from the mobile, and then changes the cell association if the received signal power is lower than the predefined threshold.
Since this reactive process cannot predict the sudden link deterioration, a mobile launches the RLF recovery procedure only after the occurrence of the LoS blockage, causing a severe degradation of the link quality.
This issue will be even more serious in 6G ultra-dense network (UDN) since the cell association occurs frequently in the cell-free network.

To support the seamless change of cell association and therefore improve the link quality, one can consider a preemptive cell association strategy that forecasts future conditions using the sequences of visual sensing data.
Specifically, by predicting the LoS/non-line-of-sight (NLoS) status between the BS and mobile, one can easily identify the set of BSs ensuring the LoS transmission.
Since the cell association decision can be made proactively before the link failure, seamless guarantee of quality of service (QoS) is possible without incurring intermediate coverage hole. 
In the predictive CV model, DL engines specialized for extracting temporal correlations (e.g., long short-term memory (LSTM) or Transformer) can be used to extract the movement pattern of a mobile from the sequential images.
Using the extracted trajectory, the CV model predicts the LoS/NLoS status, cell association, and user rate in the upcoming time frame (see Fig.~\ref{fig3}).

\subsection{Sensing and CV-aided Environment-aware Channel Estimation}
Channel acquisition is one of the most important tasks in wireless systems as the channel information is used for the symbol detection, beamforming, spatial multiplexing, interference management, resource allocation, and many more.
In 5G NR, the channel information is inferred via the received pilot signals (e.g., channel state information reference signal (CSI-RS)).
Since the amount of pilot resources needed for the channel estimation is proportional to the number of antennas, the pilot overhead issue will be a big concern in the mmWave and THz systems where hundreds to thousands of antennas are employed.

One approach to reduce the pilot overhead in the channel estimation is the neural radiance fields (NeRF), a CV technique that constructs a 3D representation of an environment using 2D images (see Fig.~\ref{fig3})~\cite{mildenhall2020nerf}.
In the NeRF-based channel estimation, multiple images taken from different angles are integrated to generate a comprehensive 3D map of the wireless propagation environments, encompassing transmitters, obstacles, reflectors, and receivers.
Using the obtained 3D wireless environments, one can easily reconstruct the geometric channel parameters such as the number of paths, angles of departure/arrival (AoDs/AoAs), and distances.
The fundamental principle behind NeRF is that 2D pixels are projections of 3D points onto the image plane so that the RGB value of each pixel in the 2D image is expressed as the weighted sum of pixels in 3D points.
By exploiting multiple 2D images captured from various angles, NeRF performs the neural network regression between the input (i.e., 3D coordinates and direction) and output (i.e., RGB values and opacity) to construct the 3D representation of the wireless environments.
Once the 3D map of wireless propagation environments is generated, one can extract the channel response by exploiting the ray tracing technique. 
In the ray tracing technique, a large number of rays are launched in a wide range of directions and then the propagation characteristics of the rays that reach the receiver are examined to extract the proper geometric channel parameters.
Since a large part of geometric channel parameters is extracted from the visual sensing information, the pilot overhead can be reduced considerably.

\subsection{Sensing and CV-aided Semantic Signal Compression}
6G will support various human-centric services such as wireless brain-computer interface (BCI), interactive hologram, and intelligent humanoid robots, which will rely more on human-related knowledge and experience-based metrics.
To this end, semantic communications that go beyond the conventional Shannon paradigm have gained attention recently~\cite{yang2022semantic}.
Main wisdom of semantic communication is to extract, transmit, and recover the information essential for the end-user task (see Fig.~\ref{fig3}).
For example, when we perform the emergency monitoring, transmission of whole raw data is unnecessary since what is needed at the end is the the core information (e.g., the nature of the emergency, location, the number of injuries, and response instructions).
Since the essential information is handful in most cases, one can significantly reduce the amount of information to be transmitted.

To convert the coarse information of an image into the semantically dense feature vector, one can use the image captioning, a technique condensing essential details of an image to a few words or sentences.
Since the size of language information (e.g., a few hundred bits) is far smaller than that of sensing image (e.g., a few megabits), transmission overhead can be reduced significantly.
For instance, in an emergency situation (e.g., fires and accidents), one can quickly recognize the event from the visual data collected from surveillance cameras, drones, and smartphones and then represent it using a descriptive sentence (e.g., ``drunk man falls a sleep at the intersection").
Moreover, to provide more specific and contextual information about the visual scene, one can use visual question and answering (VQA), a technique to comprehend the sensing data and then generate answers using a large language model (LLM) such as ChatGPT.
VQA provides the real-time answers to the questions without a complicated signal exchange so that one can promptly access the core information such as the extent of damage, potential hazards, current and next actions, and prioritize areas/actions for immediate attention.

\subsection{Sensing and CV-aided Random Access}
Random access is a process through which a mobile device can access a shared wireless channel (e.g., physical uplink shared channel (PUSCH)) to establish a connection to the network.
To distinguish mobiles during the random access process, a predefined signal called preamble is used.
In 5G NR, a mobile chooses the random access preamble from a set of preambles transmitted over SSB beams and then feeds back the chosen preamble through the physical random access channel (PRACH)~\cite{chakrapani2020design}.
Major drawback of the conventional random access is that when multiple adjacent mobiles might attempt to access the network using the same preamble in the crowded area, random access fails due to the collision.
In such cases,  mobiles should start over the random access process after the backoff period.
With the proliferation of connected devices (i.e., 10 million $\text{devices}/\text{km}^{2}$), the random access latency in 6G can easily exceed a few hundred milliseconds, violating the ultra-reliable low latency communications (URLLC) requirements.

To handle the problem, one can consider the crowd estimation, a CV technique that estimates the density of objects (e.g., human, mobile) in different regions in an image (see Fig.~\ref{fig3}).
Using the crowd density information, the BS can proactively assign a large number of preambles to the dense area.
Since mobiles in dense area choose the preambles from an expanded set of preambles, a chance of collision is reduced considerably and so will be the access latency.
In essence, the crowd estimation exploits the property that features extracted from the dense region and sparsely occupied region differ significantly.
In fact, mobiles in a dense region are predominantly depicted by the local features (e.g., hands, shoulders, heads) since they are partially obscured by other mobiles in close proximity while mobiles in a sparsely occupied region are represented by the global features (e.g., human, body).
Exploiting this property, the crowd estimation network can separate the crowded region from uncrowded one and thus generate the density map indicating the spatial distribution of mobiles.

\begin{figure}[t]
    \centering
    \includegraphics[width=1\linewidth]{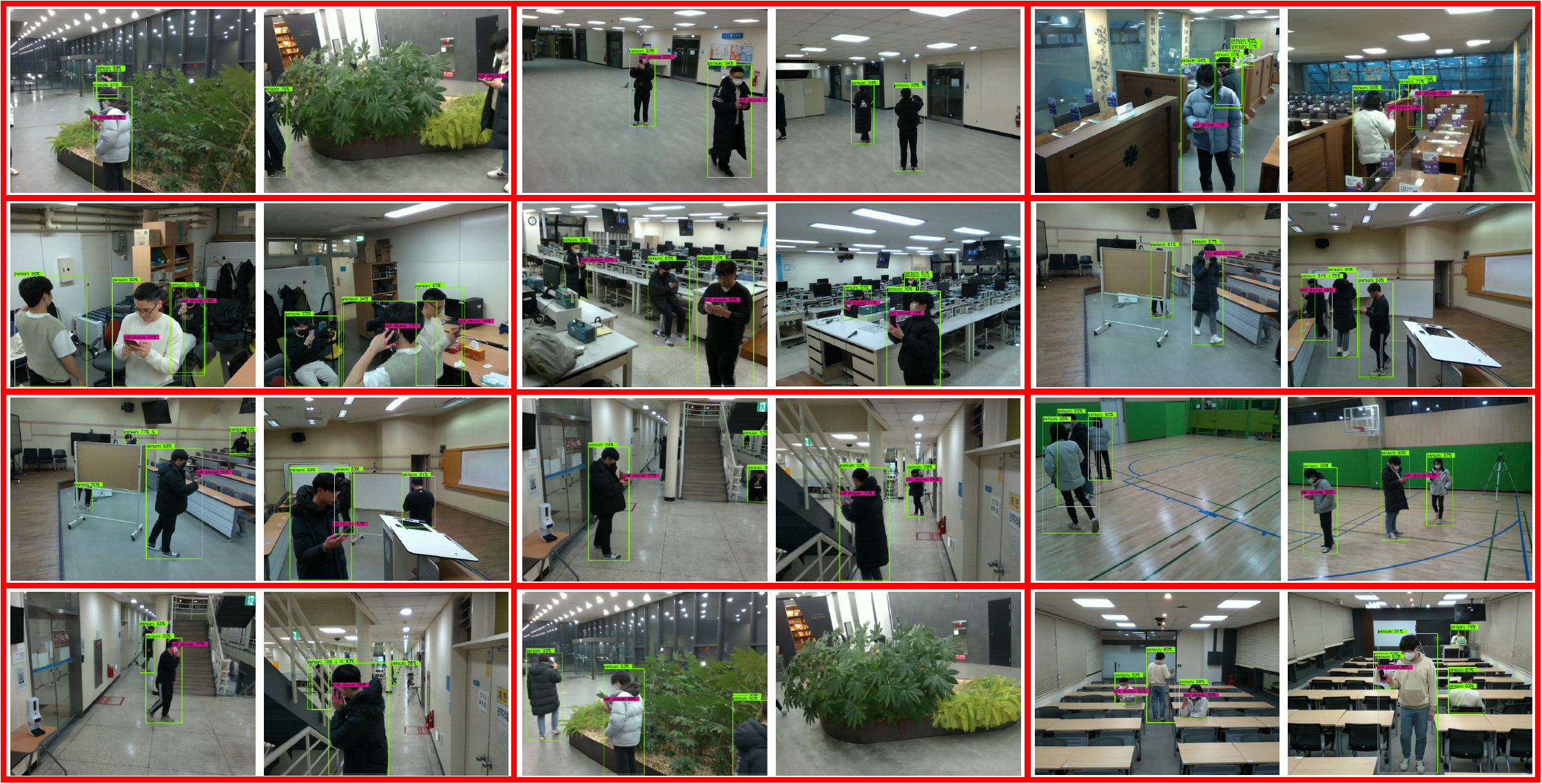}
    \caption{Example of sensing dataset tailored for wireless communications. Two cameras are used to obtain the images of a scene from different viewpoints.}
    \label{fig4}
\end{figure}

\section{Simulation Results}
To demonstrate the efficacy of the proposed SVWC framework, we conduct experiments on three key use cases: beam management, cell association, and random access.

\subsection{Sensing Dataset and Simulation Setup}
To evaluate the performance of SVWC, we generate the sensing dataset tailored for wireless communications using two Intel RealSense L515 RGB-d cameras (see Fig. \ref{fig4}).
The dataset consists of $104$ pairs of RGB and depth images acquired from $10$ distinct wireless environments including classroom, hallway, stair, gym, and food court (see https://github.com/islab-github/VOBEM2).
In each image, up to $7$ people appear, each holding a mobile within a maximum distance of $12\,$m from the cameras.
After obtaining images from two cameras, we mark the 3D bounding box around each object with a class annotation (human, mobile), distance, angle, and size on each image.
To perform the object detection and tracking in the beamforming and cell association tasks, we use DETR~\cite{carion2020end}, the state-of-the-art Transformer-based object detector.
Specifically, we use the model pre-trained on the MS-COCO 2017 dataset consisting of $80$ classes of objects and $200,000$ training images.
In the crowd estimation for the random access task, we use ConvNeXt network pre-trained on the ImageNet dataset consisting of $21,000$ classes of objects and $14$ million training images.

In our simulations, we consider THz UDN systems where $8$ BSs cooperatively serve $32$ mobiles.
As for the channel model, we use THz LoS channel model with the $0.1\,$THz carrier frequency and $100\,$MHz bandwidth and the indoor path loss model specified in 3GPP TR 38.901 Rel. 17~\cite{TR38901}.
The BSs are equipped with two RGB-d cameras and $8\times8$-uniform planar array (UPA) antennas and the mobiles are equipped with $2\times2$-UPA antennas.
To compare the positioning accuracy, we use three benchmark schemes: 1) EfficientDet-based single-view CV-aided beam management scheme~\cite{ahn2022towards}, 2) 5G positioning reference signal (PRS)-based localization scheme, and 3) 5G NR beam management scheme using $6$-bit DFT-based beam codebook~\cite{TR38855}. 
To facilitate the beam refinement using $4$ CSI-RS beams, we apply an oversampling ratio of $4$ to the beam codebook.
Also, to compare the cell association performance, we use the 5G NR cell association and two DL-based cell association schemes that perform proactive handovers by monitoring the changes of RSRP at multiple BSs~\cite{liu2020proactive, lee2022intelligent}.
Lastly, to compare the random access performance, we use two 5G NR random access scenarios where the numbers of SSBs are set to $S=8$ and $S=32$, respectively.
Note that the SSB beam sweeping period and the re-trial period are set to $20\,$ms and $10\,$ms, respectively.

\begin{table}[t]
    \centering
    \scalebox{0.91}{
    \begin{tabular}{|c|cc|cc|c|}
        \hline
        & \textbf{Human} & \textbf{Cell phone} & \multicolumn{2}{c|}{\textbf{Positioning error}} & \textbf{Normalized}\\
          & precision/recall (\%) & precision/recall (\%) & Distance (cm) & az/el angles (deg.) & \textbf{array gain} \\
         \hline
        \textbf{Multi-view SVWC (DETR)} & $96.92\,/\,98.29$    & $93.75\,/\,95.57$    & $8.17$    & $0.41\,/\,0.32$ & 0.99\\
        \textbf{Single-view SVWC (DETR)} & $96.88\,/\,85.04$    & $93.75\,/\,64.08$    & 38.50    & $2.17\,/\,2.08$ & 0.96
\\
        \textbf{Single-view SVWC (EfficientDet)} & $79.59\,/\,46.01$    & $84.71\,/\,42.09$    & $60.26$    & $3.36\,/\,3.30$ & 0.91 \\
        \textbf{5G NR PRS-based localization} & - & - & 89.67 & 3.62 / 3.63 & 0.89 \\
        \textbf{5G NR beam management} & - & - & 128.58 & $5.6\,/\,5.6$ & 0.77\\
         \hline
        \end{tabular}
        } 
    \caption{Detection probability and positioning accuracy of SVWC.}
    \label{tab1}
\end{table}

\begin{figure}[t]
    \centering
    \subfloat[]{
    \centering
    \includegraphics[width=0.482\linewidth]{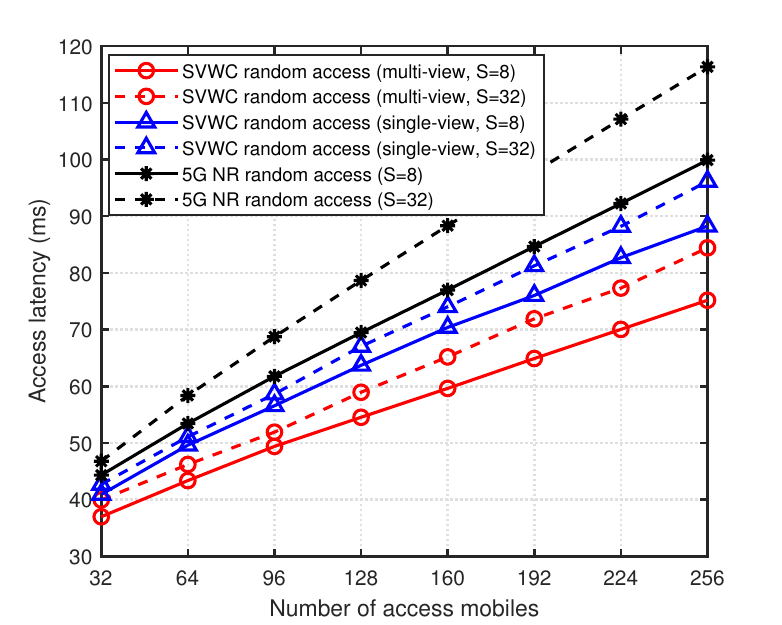}    
    }
    \hfill
    \subfloat[]{
    \centering
    \includegraphics[width=0.482\linewidth]{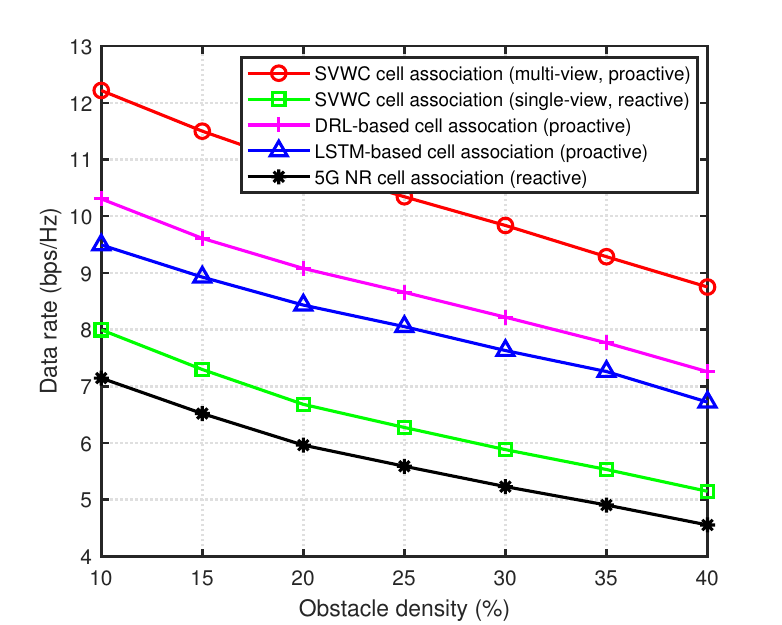}
    }
    \caption{(a) Access latency and (b) data rate performance of SVWC.}
    \label{fig5}
\end{figure}

\subsection{Simulation Results}
In Table I, we compare the beam management performances of SVWC and conventional 5G NR beam management in terms of precision and recall, positioning error, and normalized array gain.
The precision is the percentage of correctly detected objects among total detected objects and the recall is the percentage of detected objects among all target objects.
We see that the DETR model successfully identifies the class and location of a mobile in the sensing image.
Specifically, the positioning error of SVWC with a single RGB-d camera (i.e., single-view SVWC) is $8.17\,$cm on average while that of the conventional 5G NR PRS-based localization scheme is $89.67\,$cm, achieving more than $91\%$ positioning error reduction.
One can further enhance the detection probability and the positioning accuracy of SVWC by exploiting multiple images taken from different viewpoints.
We also observe that SVWC achieves a significant enhancement in the array gain over the conventional schemes.

We next evaluate the random access latency of SVWC random access scheme in Fig.~\ref{fig5} (a).
In our experiments, the access latency is defined as the sum of initial SSB beam sweeping period and re-trial period when the random access fails.
Since SVWC identifies the crowd density and allocates a large number of preambles to the SSB beams for the dense area, the number of collisions is reduced significantly and so is the access latency.
Since the 5G NR random access allocates preambles equally to all SSB beam directions, it is no wonder to see the frequent collisions and large access latency for the crowded area.
In fact, when the number of access mobiles is $256$ and the number of SSBs is $32$, the access latency of SVWC is $84\,\text{ms}$ but that of 5G NR random access scheme is $118\,\text{ms}$, achieving $29\%$ reduction in access latency. 

In Fig.~\ref{fig5} (b), we plot the data rate performance of SVWC cell association scheme as a function of the obstacle density.
The obstacle density is defined as the ratio of the area occupied by obstacles over the total service area.
We observe that SVWC cell association scheme achieves $84\%$ and $19.6\%$ data rate improvements over the 5G NR cell association and DL-based schemes.
As mentioned, SVWC predicts the LoS/NLoS status and user rate and then performs the preemptive handover so that it ensures a seamless LoS transmission and reliable data transmission without the link failure.
Whereas, the conventional cell association schemes suffer from a severe link deterioration due to the reactive handover after the occurrence of the LoS blockage.

\section{Future Works and Conclusion}
In this article, we outlined a sensing and computer vision-aided wireless communication (SVWC) framework for 6G systems and examined its feasibility through realistic wireless tasks including beam management, cell association, channel estimation, semantic signal compression, and random access.
Using the specially designed vision dataset tailored for wireless communications, we demonstrated the efficacy of SVWC in terms of positioning accuracy, data rate, and access latency.
Considering the immense potential of visual sensing and CV techniques, we are confident that SVWC will serve as one of the key pillars in 6G.
As we embark on the initial stage of new paradigm, there are many interesting open questions that remain unanswered.
We here outline some future research directions.

\begin{itemize}
\item \textbf{Seamless coverage provision via multi-modal sensing}: 
One of the major challenges of SVWC is to ensure seamless service quality even in the demanding scenarios with obstacles, blind spots, or low light condition (e.g., night-time and rainy weather).
To address this issue, one can exploit a multi-modal sensing, which utilizes multiple sensing modalities simultaneously.
For instance, in the NLoS scenario where a mobile is visually blocked by obstacles, multi-view cameras with different orientations or sensors employing relatively long-wavelength light (e.g., ultrasonic sensor and radar) can be used to detect the mobiles blocked by the obstacles.
Also, in low light environments, integration of the RGB camera and invisible light sensors (e.g., infrared camera and thermographic camera) can substantially enhance the positioning accuracy of SVWC.

\item \textbf{Wireless environment compatible training}:
Yet another key challenge of SVWC is to assure the environment compatibility.
To support various wireless environments, one should train SVWC model using the data collected from various communication environments.
To reduce the overhead, one can consider the transfer learning, an approach to pre-train the DL model using a large amount of general datasets and then fine-tune it using a few datasets tailored for the target environment.
As long as the pre-trained SVWC model recognizing the common visual characteristics is obtained, one can generate a fine-tuned SVWC model for the new environment with just a handful of data.

\item \textbf{Privacy-preserving DL model design}:
Since SVWC exploits visual sensing information containing human and mobile, preserving privacy is of great importance.
In a nutshell, there are several approaches to deal with this issue.
One approach is to use the low-resolution visual sensing mechanism in the object detection.
Essence of this approach is to capture the low-resolution sensing information (e.g., blurred image) that does not contain personal appearance details.
Then,	 by using the super-resolution technique, the BS enhances the resolution of a part of image that is necessary for the object detection (e.g., edges, contours, hands, clothes).
Another approach is to use the privacy-preserving object detection technique that performs the object detection for the specified objects.
Yet another approach is to use non-camera sensing modality (e.g., LiDAR 3D point cloud data, radar image, and heat map) that captures only the shape of an object, not the detailed appearance of a person.

\item \textbf{Energy-efficient and low-latency vision processor design}:
Two natural concerns when implementing the DL technique for wireless systems are considerable energy consumption and DL processing latency.
At present, the power consumption of \textit{Qualcomm Snapdragon 888}, the latest artificial intelligence (AI)-focused processor, and Intel RealSense L515 RGB-d camera is $5\,$W and $4\,$W and the SSB beam transmission power is $20\,$W.
Also, the frame rate of DETR is $28$ frame per second (fps), which implies that the processing latency for each image is roughly $\frac{1}{28}\approx 35\,$ms~\cite{carion2020end}.
The processing latency decreases to $8.7\,$ms when the real time (RT)-DETR, a streamlined version of DETR, is used.
When compared to this, the beam sweeping latency of 5G NR is $20\,$ms so that SVWC can achieve more than $56\%$ latency reduction over the conventional 5G NR beam management scheme.
Considering that there will be various on-going efforts for the fast and low-power AI processors with a few nano-scale CMOS technology, we expect that the power consumption and processing latency of SVWC can be further reduced down the road.
\end{itemize}

\section*{Acknowledgements}
This work was supported by the National Research Foundation of Korea (NRF) grant funded by the Korea government (MSIT) (2022R1A5A1027646, RS-2023-00208985, and RS-2023-00252789).

\bibliographystyle{IEEEtran}

\begin{IEEEbiography}[{\includegraphics[width=1in,height=1.25in,clip,keepaspectratio]{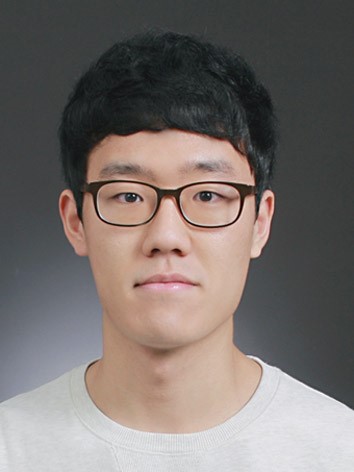}}]{Seungnyun Kim} (snkim94@mit.edu)
received the B.S. degree from the Department of Electrical and Computer Engineering, Seoul National University (SNU), Seoul, Korea, in 2016, where he received the M.S. and Ph.D. degrees in 2023.
He is currently a postdoctoral researcher with the Wireless Information and Network Sciences Laboratory, Massachusetts Institute of Technology (MIT), Massachusetts, USA. 
His research interests include beyond 5G and 6G wireless communications, optimization, and machine learning.
\end{IEEEbiography}

\begin{IEEEbiography}[{\includegraphics[width=1in,height=1.25in,clip,keepaspectratio]{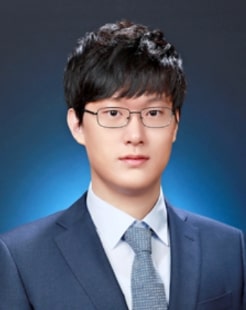}}]{Jihoon Moon} (jhmoon@islab.snu.ac.kr)
received the B.S. degree from the Department of Electrical and Computer Engineering, SNU, Seoul, Korea, in 2019, where he is working toward the Ph.D. degree.
His research interests include sensing and deep learning techniques for 6G.
\end{IEEEbiography}

\begin{IEEEbiography}[{\includegraphics[width=1in,height=1.25in,clip,keepaspectratio]{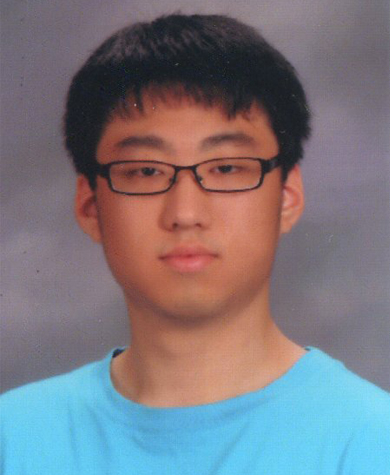}}]{Jinhong Kim} (jhkim@islab.snu.ac.kr)
received the B.S. degree from the Department of Electrical and Information Engineering, Seoul National University of Science and Technology, Seoul, Korea, in 2016 and the Ph.D. degree in Electrical and Computer Engineering, SNU, Seoul, Korea, in 2023. 
Since September 2023, he has been with Samsung Electronics, where he is working for 5G and 6G modem design. 
His research interests include sparse signal recovery and deep learning techniques for the 6G wireless communications.
\end{IEEEbiography}

\begin{IEEEbiography}[{\includegraphics[width=1in,height=1.25in,clip,keepaspectratio]{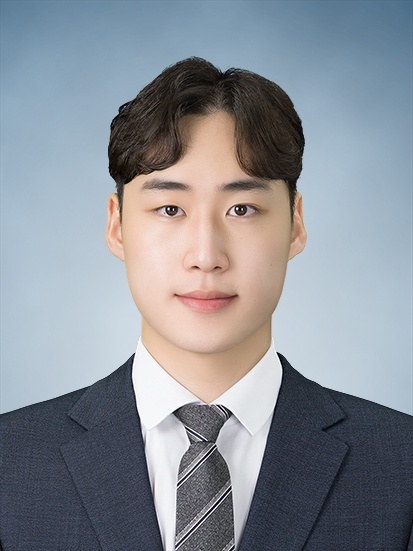}}]{Yongjun Ahn} (yjahn@islab.snu.ac.kr)
received the B.S. degree from the Department of Electrical and Computer Engineering, Seoul National University (SNU), Seoul, Korea, in 2019, where he received the M.S. and Ph.D. degrees in 2024.
Since March 2024, he has been with Samsung Electronics, where he is working for the development of 6G communication technologies. 
His research interests include 6G wireless communications and deep learning techniques.
\end{IEEEbiography}

\begin{IEEEbiography}[{\includegraphics[width=1in,height=1.25in,clip,keepaspectratio]{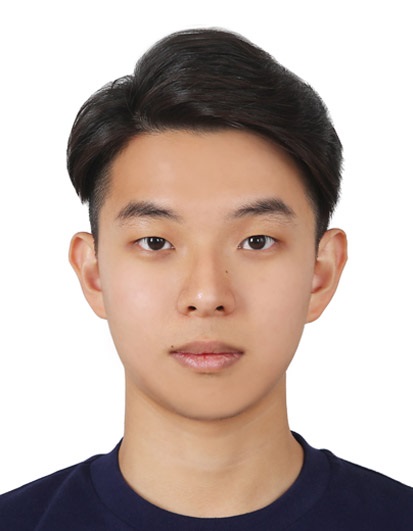}}]{Donghoon Kim} (dhkim@islab.snu.ac.kr)
received his B.S. and M.S. degree in Electronics Engineering from Kyung Hee University, Korea, in 2019 and 2021, respectively.
He is currently pursuing the Ph.D. degree in Electrical and Computer Engineering, Seoul National University. 
His research interests include deep learning algorithm design and implementations.
\end{IEEEbiography}

\begin{IEEEbiography}[{\includegraphics[width=1in,height=1.25in,clip,keepaspectratio]{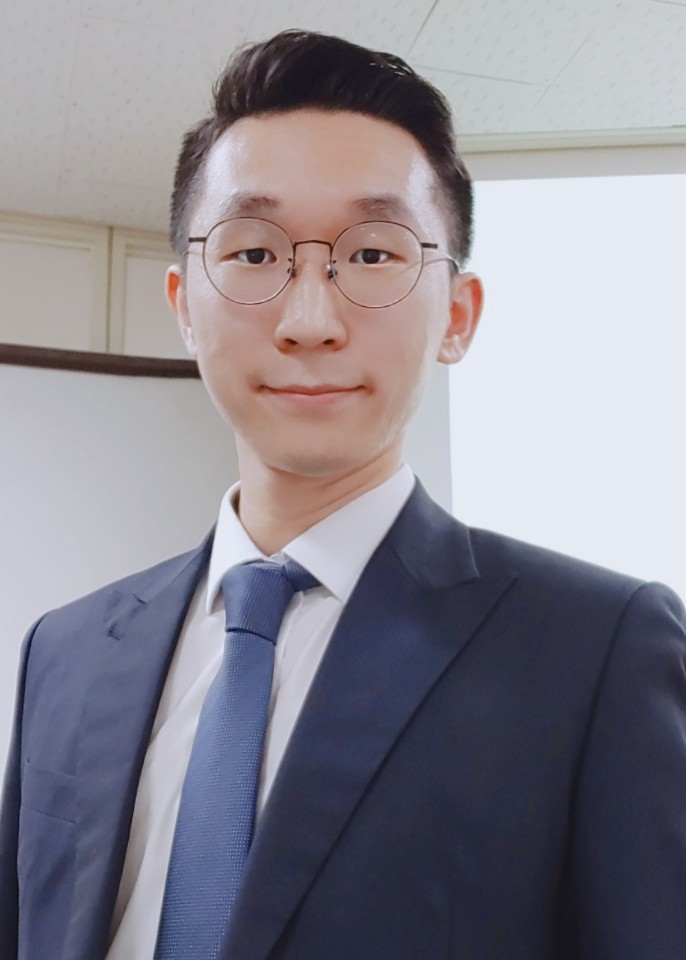}}]{Sunwoo Kim} (swkim@islab.snu.ac.kr)
received the B.S. degree in electrical engineering, Stony Brook University, New York, USA, in 2018. 
He is currently working toward the Ph.D. degree in Electrical and Computer Engineering, SNU, Seoul, Korea.
His research interests include deep learning algorithm design and implementations.
\end{IEEEbiography}

\begin{IEEEbiography}[{\includegraphics[width=1in,height=1.25in,clip,keepaspectratio]{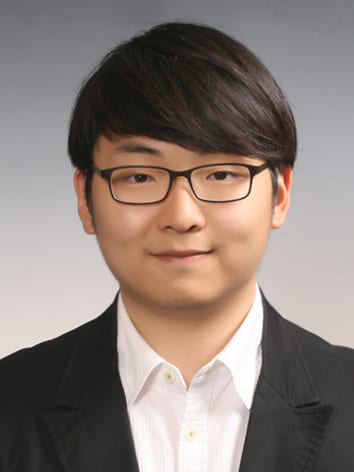}}]{Kyuhong Shim} (khshim@islab.snu.ac.kr)
received the B.S. degree from the Department of Electrical and Computer Engineering, SNU, Seoul, Korea, in 2015, where he received the Ph.D. degree in 2022. 
Since September 2022, he has been with Qualcomm Inc. where he is working for speech and language processing. 
His research interests include deep learning algorithm design and implementations.
\end{IEEEbiography}

\begin{IEEEbiography}[{\includegraphics[width=1in,height=1.25in,clip,keepaspectratio]{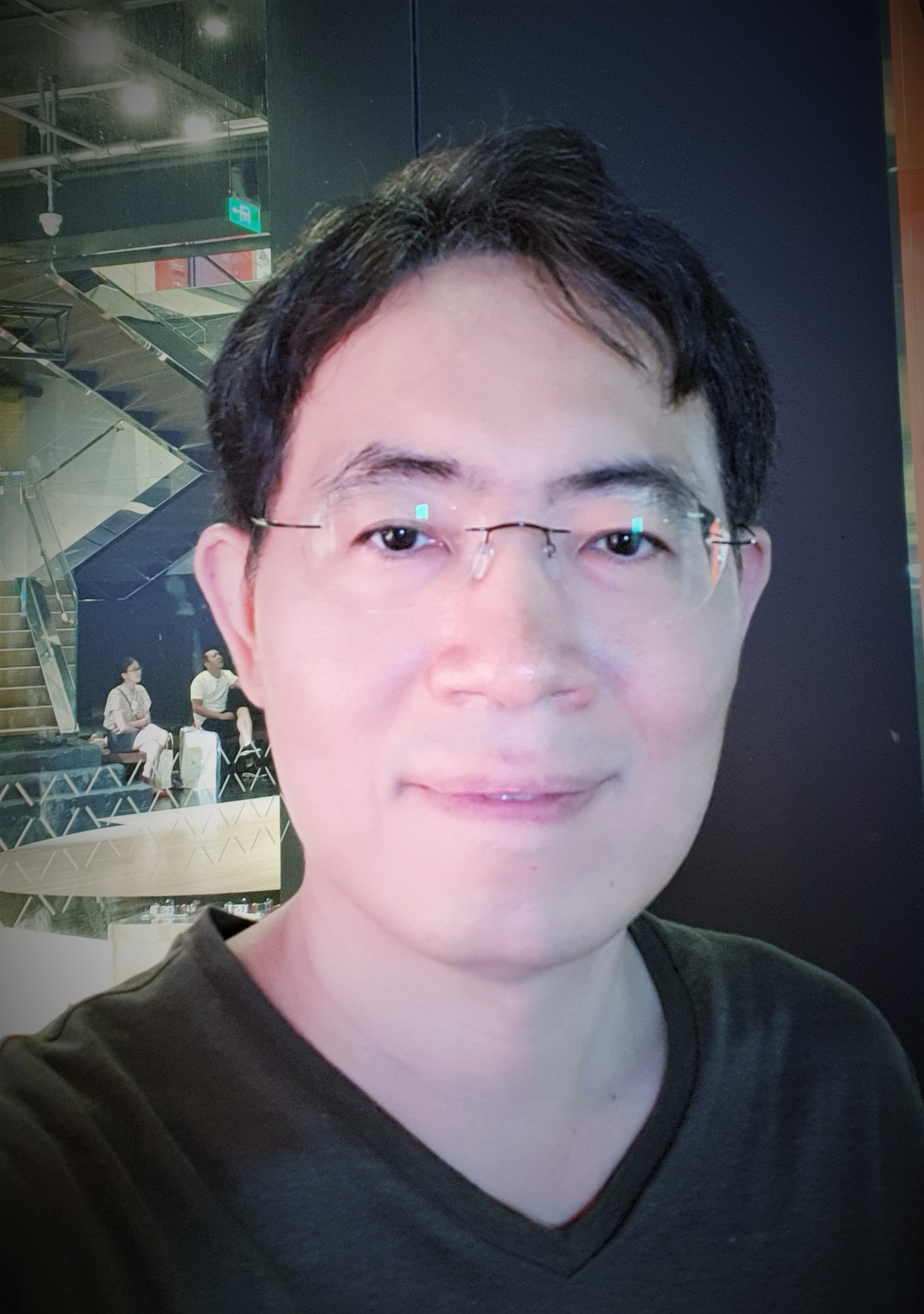}}]{Byonghyo Shim} (bshim@snu.ac.kr)
received the B.S. and M.S. degree in Control and Instrumentation Engineering (currently Electrical Eng.) from SNU, Seoul, Korea, in 1995 and 1997, respectively, and the M.S. degree in Mathematics and the Ph.D. degree in Electrical and Computer Engineering from the University of Illinois at Urbana-Champaign (UIUC), Urbana, in 2004 and 2005, respectively. 
He is currently a professor in the Department of Electrical and Computer Engineering, SNU. 
His research interests include signal processing for wireless communications, machine learning, and information theory. 
\end{IEEEbiography}

\end{document}